\documentclass[a4paper]{article}

\usepackage{INTERSPEECH2015}

\usepackage{graphicx}
\usepackage{amssymb,amsmath,bm}
\usepackage{textcomp}
\usepackage{url}
\usepackage[justification=justified,singlelinecheck=false]{subcaption}
\usepackage[]{algorithm2e}

\ninept

\title{Multi-Modal Hybrid Deep Neural Network for Speech Enhancement}

\makeatletter
\def\name#1{\gdef\@name{#1\\}}
\makeatother \name{{\em Zhenzhou Wu$^1$, Sunil Sivadas$^2$,  Yong Kiam Tan$^1$, Ma Bin$^2$, Rick Siow Mong Goh$^1$}}

\address{$^1$ Distributed Computing (DC Group), Institute of High Performance Computing, Singapore. \\
$^2$Human Language Technology Department, Institute for Infocomm Research (I$^2$R), Singapore.\\
  {\small \tt \{wuzz,tanyk,gohsm\}@ihpc.a-star.edu.sg \{sivadass,mabin\}@i2r.a-star.edu.sg} 
}

\begin{document}

\maketitle
\begin{abstract}
Deep Neural Networks (DNN) have been successful in enhancing noisy speech signals. Enhancement is achieved by learning a nonlinear mapping function from the features of the corrupted speech signal to that of the reference clean speech signal. The quality of predicted features can be improved by providing additional side channel information that is robust to noise, such as visual cues.  In this paper we propose a novel deep learning model inspired by insights from human audio visual perception. In the proposed unified hybrid architecture, features from a Convolution Neural Network (CNN) that processes the visual cues and features from a fully connected DNN that processes the audio signal are integrated using a Bidirectional Long Short-Term Memory (BiLSTM) network. The parameters of the hybrid model are jointly learned using backpropagation. We compare the quality of enhanced speech from the hybrid models with those from traditional DNN and BiLSTM models.
\\
\end{abstract}

\section{Introduction}
Humans integrate cues from multiple sensory organs, such as our ears and eyes, for reliable perception of real world data. When data from one of the sensory organs, such as the ear, is  corrupted by noise, the human brain uses other senses, such as sight, to reduce the uncertainty. In conversational interfaces, speech is the primary mode of communication, with visual cues augmenting the information exchange. The McGurk effect \cite{mcgurk1976hearing} is one example in speech perception where humans integrate audio and visual cues. Visual cues typically provide information about place of articulation \cite{summerfield1992lipreading} and lip shapes that aid in discriminating phonemes with similar acoustic characteristics. In acoustically noisy environments, the visual cues help in disambiguating the target speaker from the surrounding audio sources.

There are various computational models of multi-modal information fusion \cite{potamianos2004audio} for audio-visual speech processing. Deep learning provides an elegant framework for designing data driven models for multi-modal and cross-modal feature learning \cite{ngiam2011multimodal,srivastava2012multimodal}. In \cite{ngiam2011multimodal}, stacks of Restricted Boltzmann Machines (RBMs) \cite{hinton2006reducing} were trained to learn joint representations between acoustic features of phonemes and images of the mouth region. Their bimodal deep autoencoder with shared hidden layer representation was able to capture the higher level correlation between acoustic features and visual cues. 

In all of the above representations, features of both modalities are learned through fully connected DNNs. Thus, these models are homogeneous in their architecture even though their input modalities are heterogeneous. It is well known that human visual processing is better modeled by CNNs \cite{krizhevsky2012imagenet}. Higher level feature processing in the human brain also typically involves units that model long term dependencies among lower level features. Deep learning models with memory cells, such as LSTM \cite{hochreiter1997long} and BiLSTM \cite{graves2005framewise} networks, have out-performed fully connected DNNs and CNNs in noise robust speech recognition \cite{graves2013speech,maas2012recurrent}. In this paper we propose a novel hybrid deep learning architecture where the acoustic features are first extracted by a fully connected DNN and the visual cues by a CNN. Higher level long-term dependencies among these auditory and visual features are modeled by a BiLSTM. The parameters of this multi-modal hybrid network are jointly optimized using backpropagation. The models are validated on an artificially corrupted audio-visual database \cite{LDC2009V01}.

In the following sections, we present a brief background of existing multi-modal and hybrid deep learning models (Section~\ref{prior}). Subsequently, the architecture of the proposed hybrid model is presented in detail (Section~\ref{hybrid}). Finally, we report experimental details (Sections~\ref{dnn-lstm},\ref{exp}) and conclude (Section~\ref{conclusion}).

\section{Related Work}\label{prior}
Recently, there has been increased interest in heterogeneous deep learning architectures \cite{sainath2015convolutional,deng2014ensemble}. These architectures combine the strengths of constituent deep learning models to learn better high level abstractions of features. In \cite{deng2014ensemble}, an ensemble model for phoneme recognition was proposed where a CNN and RNN were first independently trained to compute ``low-level" features. A linear ensemble model was then trained to combine the posterior probabilities from these lower level classifiers. This model followed the strategy of stacking classifiers to achieve better discrimination and generalization \cite{wolpert1992stacked}. In \cite{sainath2015convolutional}, the model combines CNNs, LSTMs and DNNs into a unified framework. Firstly, a CNN was used to reduce spectral variability and its output features were then fed into a LSTM to reduce temporal variability. Finally, the output of the LSTM is processed by a DNN and the whole model is trained jointly. The multi-modal deep learning model proposed in \cite{ngiam2011multimodal} used sparse RBMs for combining the different lower level modalities. The model we propose combines the strengths of the above models. Our model has a fully connected DNN that takes a few frames of acoustic features as input, and an image processing CNN model that computes a higher level image representation of the lip movements over the same window. The features from these models are concatenated to form a shared representation, which is fed into a BiLSTM model to capture the temporal and spatial inter-dependencies between the audio and visual features. We train the entire model jointly to reconstruct cleaned spectral features. We call this model a BiModal-BiLSTM. The next section explains the proposed model in more detail.

\begin{figure*}[t]
\begin{center}
\includegraphics[width=10cm]{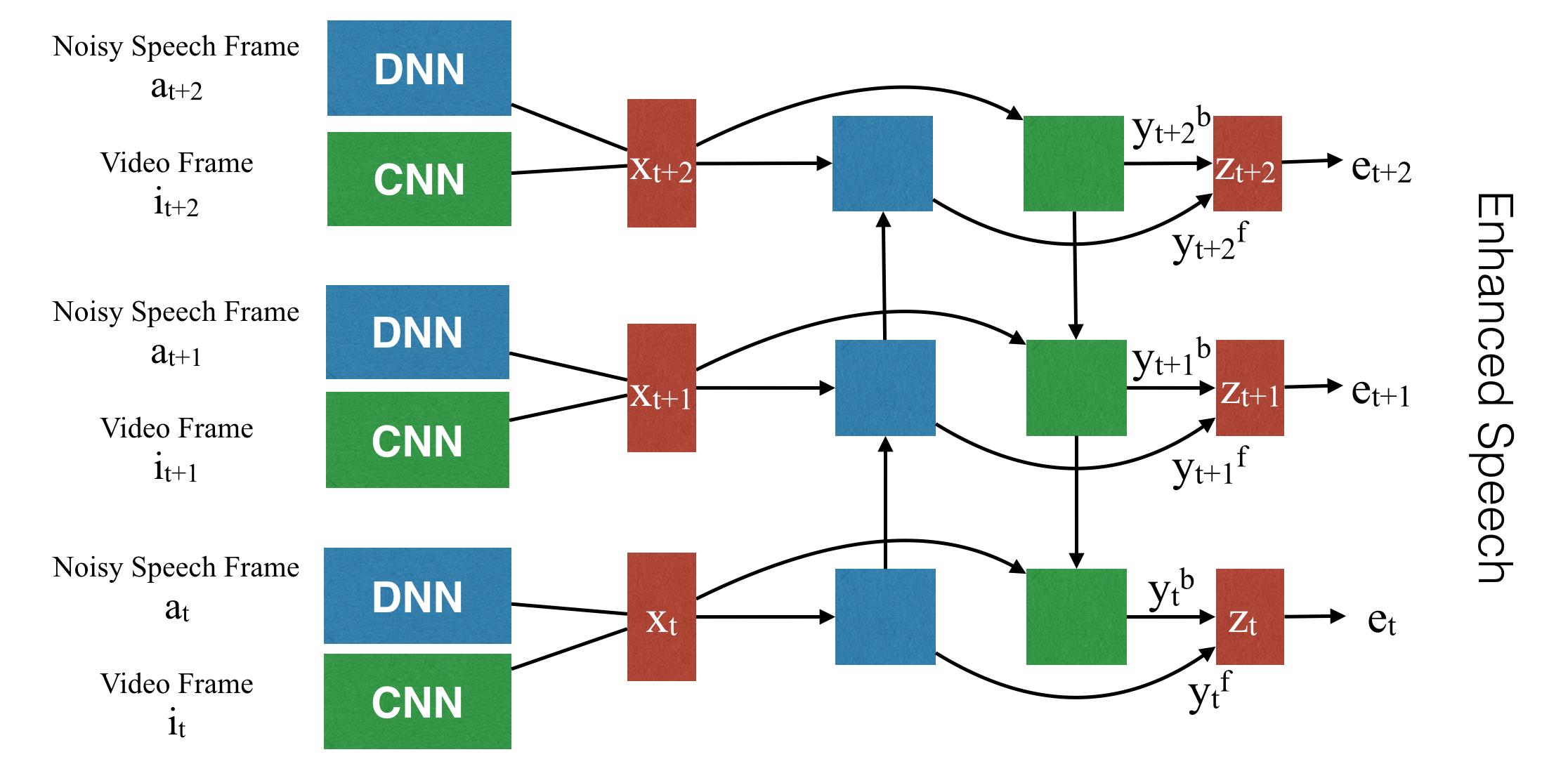}
\end{center}
\vspace{-0.5cm}
\caption{Architecture of the multi-modal hybrid deep neural network.}
\vspace{-0.1cm}
\label{fig:two_channel_rnn}
\vspace{-0.1cm}
\end{figure*}

\section{BiModal-BiLSTM Model}\label{hybrid}
In the BiModal-BiLSTM model, we take in an image channel \ensuremath{i_t} and an audio channel \ensuremath{a_t} at each time-step. For the image channel, we use a CNN to extract a high level feature representation 
\begin{equation}
\label{eqn:cnn}
i^{*}_{t} = \text{CNN}(i_t)
\end{equation}
and for the audio channel, we use a DNN to transform the audio features into a learned representation at the upper layer of the DNN. 
\begin{equation}
\label{eqn:dnn}
a^{*}_{t} = \text{DNN}(a_t)
\end{equation}
Then, we concatenate the two features $x_{t} = \text{Concat}(i^{*}_{t}, a^{*}_{t})$ and pass the joint representation into a BiLSTM model which consists of a forward LSTM, 
\begin{equation}
    y^{f}_{t}, h_{t+1} = \text{FLSTM}(x_{t}, h_{t})
\end{equation}
 and a backward LSTM, 
 \begin{equation}
 y^{b}_{t}, h_{t-1} = \text{BLSTM}(x_{t}, h_{t})
 \end{equation}
The FLSTM and BLSTM are standard LSTM models, as defined in \cite{graves2005framewise,Hochreiter:1997} except that they unroll in opposite time direction. The concatenated feature $x_t$ contains bimodal information from audio and image. The output feature from FLSTM $y_t^{f}$ contains information from the past frames and the BLSTM output feature $y_t^{b}$ contains information from the future. Therefore when we sum these two features $z_t = y_t^{f} + y_t^{b}$ and use it to reconstruct the enhanced speech frame $e_t = \text{FC}(z_t)$ with a fully connected layer, the enhanced speech frame $e_t$ will have access to bidirectional information for the past and future from both input audio and image channels which helps in speech enhancement. Figure~\ref{fig:two_channel_rnn} shows the schematic of the hybrid model.



\section{Baseline Models}\label{dnn-lstm}

To understand the effectiveness of BiModal-BiLSTM model, we designed two baseline models with similar number of parameters to answer two questions:
\begin{enumerate}
\item Does having an additional image modality help in model generalization for speech enhancement?
\item Does the BiLSTM work better than a purely feed-forward neural network?
\end{enumerate}
The second question has already been answered in speech recognition and speech enhancement \cite{graves2013speech,maas2012recurrent} on speech datasets, but it will be interesting to compare the models alongside our BiModal-BiLSTM model.

\subsection{Single-Channel-BiLSTM}
The Single-Channel-BiLSTM has the same architecture as our BiModal-BiLSTM model, except that we removed the CNN image feature extractor (Equation \ref{eqn:cnn}), and only use the noisy audio channel as input. Everything else is kept the same to ensure that any difference in the final generalization result is due to the CNN image feature extractor. 

\subsection{Single-Channel-DNN}
In the Single-Channel-DNN, we take the noisy audio as input and enhance it directly with a DNN \cite{xu2015regression}. The single-Channel-DNN has the same DNN architecture as the BiModal-BiLSTM and Single-Channel-BiLSTM (Equation \ref{eqn:dnn}). However, to ensure that the total number of parameters in Single-Channel-DNN matches that of Single-Channel-BiLSTM, we appended two extra fully connected layers, so that differences in the final generalization result is due to the difference in network architecture, rather than different number of parameters.


\begin{figure*}[t]
\centering
\begin{subfigure}{0.3\textwidth}
\captionsetup{justification=centering,margin=0.1cm}
  \centering
  \includegraphics[width=1.0\textwidth]{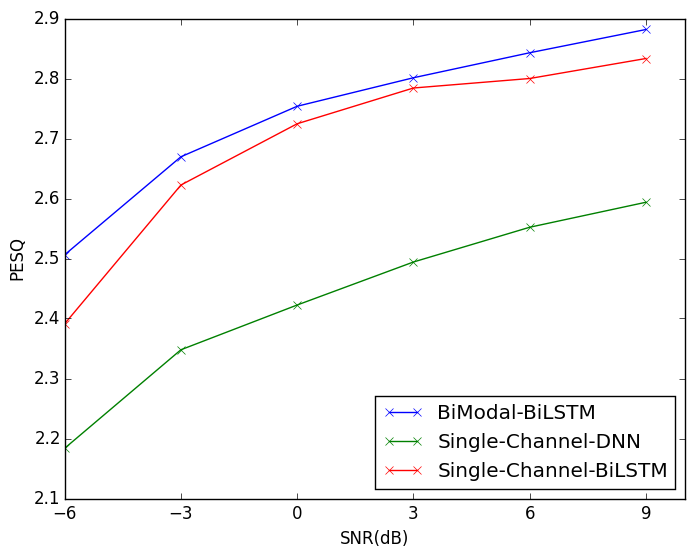}
  \caption{Alarm (Seen Noise)}
  \label{fig:alarm_pesq}
\end{subfigure}%
\begin{subfigure}{0.3\textwidth}
\captionsetup{justification=centering,margin=0.1cm}
  \centering
  \includegraphics[width=1.0\textwidth]{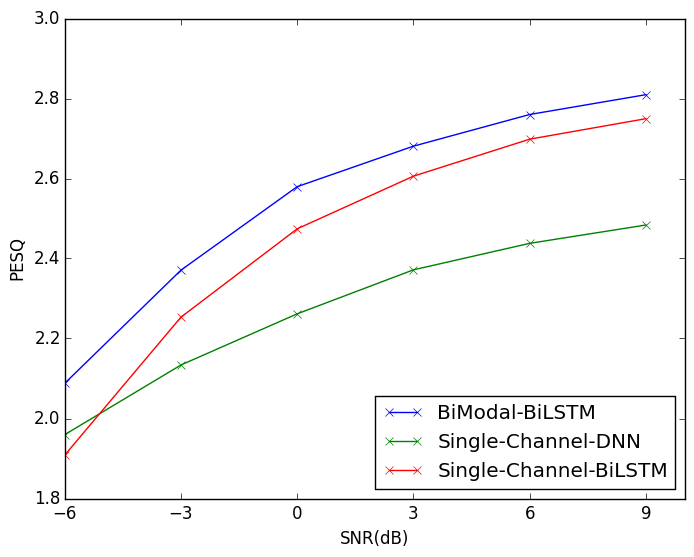}
  \caption{Crowd (Seen Noise)}
  \label{fig:crowd_pesq}
\end{subfigure}%
\begin{subfigure}{0.3\textwidth}
\captionsetup{justification=centering,margin=0.1cm}
  \centering
  \includegraphics[width=1.0\textwidth]{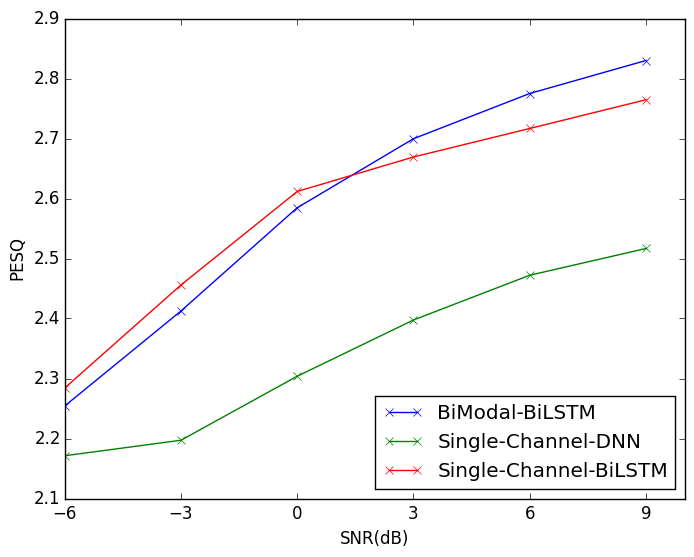}
  \caption{Traffic (Unseen Noise)}
  \label{fig:traffic_pesq}
\end{subfigure}
\vspace{-0.2cm}
\caption{PESQ scores of denoised speech generated using different models on seen and unseen noise conditions.}
\label{fig:pesq_scores}
\end{figure*}

\section{Experimental Details}\label{exp}

\subsection{Experimental Data}\label{data}

We conducted our experiments on an audiovisual dataset consisting of 14 native American English speakers~\cite{LDC2009V01}. There are 94 recorded files for each speaker, ranging from short single word clips to long recordings of multiple full sentences. We extracted nonspeech, environmental noises from an on-line corpus~\cite{noisenonspeech}\footnote{All noise samples in the same category were concatenated.}.

For our test set, we used two of the longer audio files (CID Sentences List A and NU Auditory Test No.6~List I) for each speaker. Other samples in the dataset were used to construct the training set. We corrupted each sample with each of the noise types at a selected Signal-to-Noise Ratio (SNR)\footnote{We start corrupting using a randomly selected point in each noise clip and we repeat the noise clips if they are too short.}. For the training samples, we randomly selected an integral SNR in the range [-5,5]. In total, this gave us roughly 20.7 hours of stereo training data. For the test data, we corrupted with SNRs in steps of 3 in the range [-6,9]. The training noise types were: alarm, animal, crowd, water and water; traffic noise was only used for (unseen) testing.

We extracted the log power spectrum from the audio component of each sample using a 320-point STFT with 0.02s window and 0.01s overlap. For the input to our network, we further extracted the first and second temporal derivatives for each frame and then reduced the number of dimensions to 100 using Principal Component Analysis (PCA). For the models that use visual inputs, we manually took a 100 by 160 crop around the mouth region of each speaker and further down-sample the crop to 64 by 64 for training.

Our models are trained to recover the log power spectrum of the clean audio samples from the corrupted input samples. To complete the reconstruction, we perform an inverse STFT using the recovered power spectrum together with the phase spectrum of the corrupted input. All data manipulation was done using off the shelf packages~\cite{librosa,scikit-learn}.

\begin{table}[h]
\begin{center}
\begin{tabular}{ |c|c|c|c| } 
 \hline
  & \bf{Kernel} & \bf{Stride} & \bf{Number of Filters} \\ 
  \hline
 conv1 & 5x5 & 1x1 & 8 \\ 
 pool1 & 5x5 & 2x2 & \\ 
 conv2 & 3x3 & 1x1 & 16 \\ 
 pool2 & 3x3 & 2x2 & \\ 
 conv3 & 3x3 & 1x1 & 32 \\ 
 pool3 & 3x3 & 2x2 & \\ 
 fc1 & & & 500 \\ 
 fc2 & & & 300 \\ 
 fc3 & & & output dim\\ 
 \hline
\end{tabular}
\end{center}
\vspace{-0.6cm}
\caption{CNN image feature extractor}
\label{table:cnn}
\end{table}

\vspace{-0.5cm}

\begin{table}[h]
\begin{center}
\begin{tabular}{ |c|c| } 
 \hline
  {Model} & {No. of parameters}  \\ 
  \hline
 Single-Channel-DNN & 1.80 million  \\ 
 Single-Channel-BiLSTM & 1.03 million \\ 
 BiModal-BiLSTM &  1.38 million \\ 
 \hline
\end{tabular}
\end{center}
\vspace{-0.6cm}
\caption{Number of Model Parameters}
\label{table:param}
\end{table}

\vspace{-0.5cm}

\begin{table}[h]
\begin{center}
\begin{tabular}{ |c|c|c|c| } 
 \hline
  {Model} & {Mean-Squared-Error} \\ 
  \hline
 \bf{BiModal-BiLSTM} & \bf{0.233}  \\ 
 \hline
 Single-Channel-BiLSTM & 0.243 \\ 
 \hline
 Single-Channel-DNN & 0.282  \\ 
 \hline
\end{tabular}
\end{center}
\vspace{-0.6cm}
\caption{Mean-Squared-Error on cross-validation set}
\label{table:cv-mse}
\end{table}

\vspace{-0.5cm}

\begin{table}[h]
\begin{center}
\begin{tabular}{ |c|c|c|c| } 
 \hline
  {Model} & {Alarm} & {Crowd} & {Traffic} \\ 
  \hline
 \bf{BiModal-BiLSTM} & \bf{2.74} & \bf{2.55} & \bf{2.59}  \\ 
 \hline
 Single-Channel-BiLSTM & 2.69 & 2.45 & 2.58\\ 
 \hline
 Single-Channel-DNN & 2.43 & 2.27 & 2.34  \\ 
 \hline
\end{tabular}
\end{center}
\vspace{-0.6cm}
\caption{Mean PESQ score over all SNRs for various noise conditions.}
\label{table:mean_pesq}
\end{table}


\subsection{Model Specification}
 In order to ensure a fair comparison, we chose model sizes such that they have roughly the same number of parameters. Table \ref{table:param} shows the number of parameters for each model. The DNN audio feature extractor in Equation \ref{eqn:dnn} has architecture 100\ensuremath{n}-500-300-\ensuremath{outdim} where 100 is the PCA dimension for 1 frame, \ensuremath{n} is the number of frames stacked together and \ensuremath{outdim} is the dimensionality of \ensuremath{a^{*}_{t}}. We set \ensuremath{outdim} to 350 for BiModal-BiLSTM and 400 for Single-Channel-BiLSTM and Single-Channel-DNN. Table \ref{table:cnn} shows the specifications of CNN image feature extractor from Equation \ref{eqn:cnn}. The Single-Channel-DNN consists of DNN audio feature extractor and two hidden layers of dimensions 1000-500. The Single-Channel-BiLSTM also has the DNN audio feature extractor, followed by one BiLSTM layer of 400 input dimension and 200 output dimension, and a fully connected layer of 200. The BiModal-BiLSTM has the same audio architecture as the Single-Channel-BiLSTM, but with an additional CNN image feature extractor depicted in Figure \ref{fig:two_channel_rnn}. Since we expect that the audio component contains much more information about the speech than the lip movements from the image, we bias the concatenated shared representation \ensuremath{x_{t}} to have 350 dimensions from the audio DNN, \ensuremath{a^{*}_{t}}, but only 50 dimensions from the image CNN, \ensuremath{i^{*}_{t}}.
 In all the fully-connected and convolutional layers, we used batch normalization \cite{ioffe2015batch} to reduce the internal covariate shift of the outputs from one layer to another. From our experiments, we found that this ensures stable convergence. 

\begin{figure}[!ht]
  \centering
  \begin{subfigure}{\linewidth}
    \captionsetup{justification=centering,margin=0.1cm}
    \includegraphics[width=9cm,height=2.8cm,trim=80 60 0 0,clip=true]{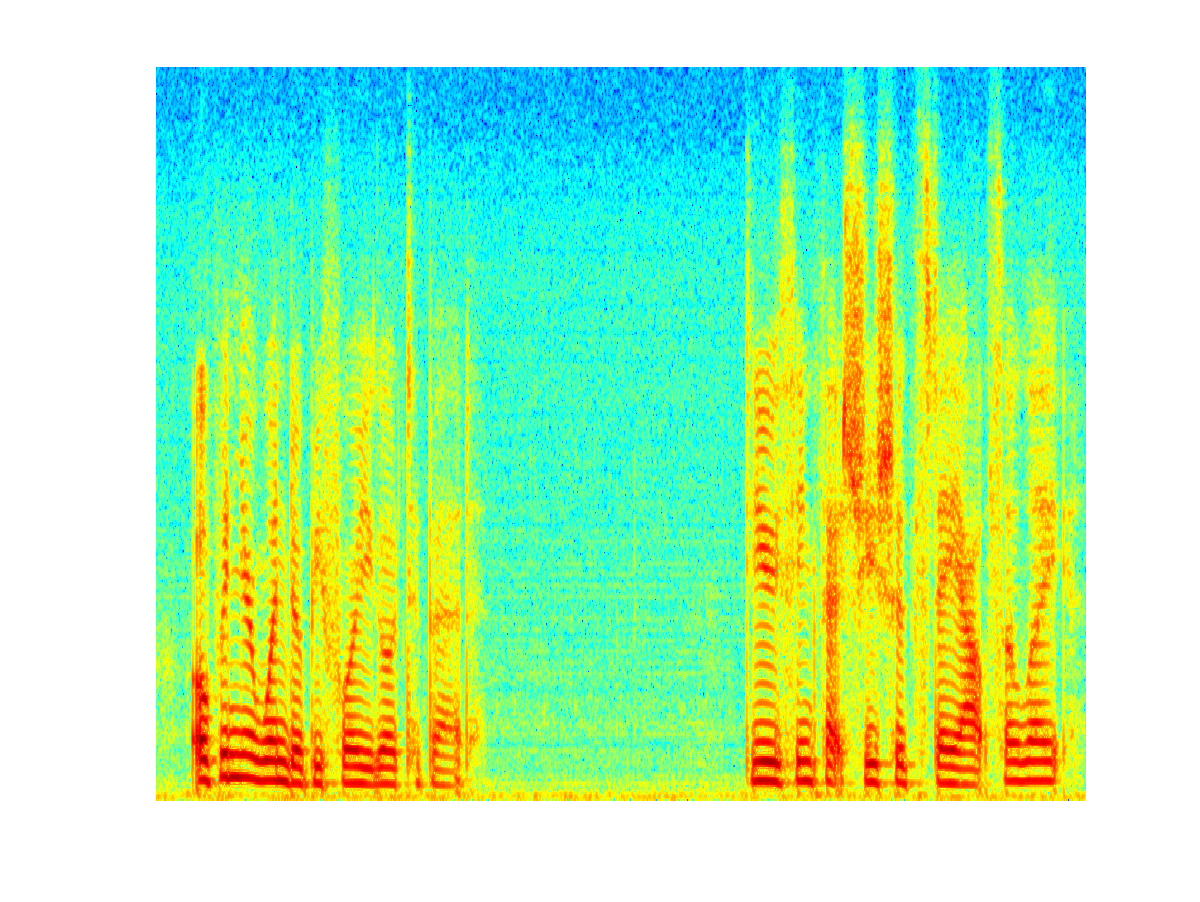}
    \caption{Clean signal}
  \end{subfigure}

  \begin{subfigure}{\linewidth}
    \captionsetup{justification=centering,margin=0.1cm}
    \includegraphics[width=9cm,height=2.8cm,trim=80 60 0 0,clip=true]{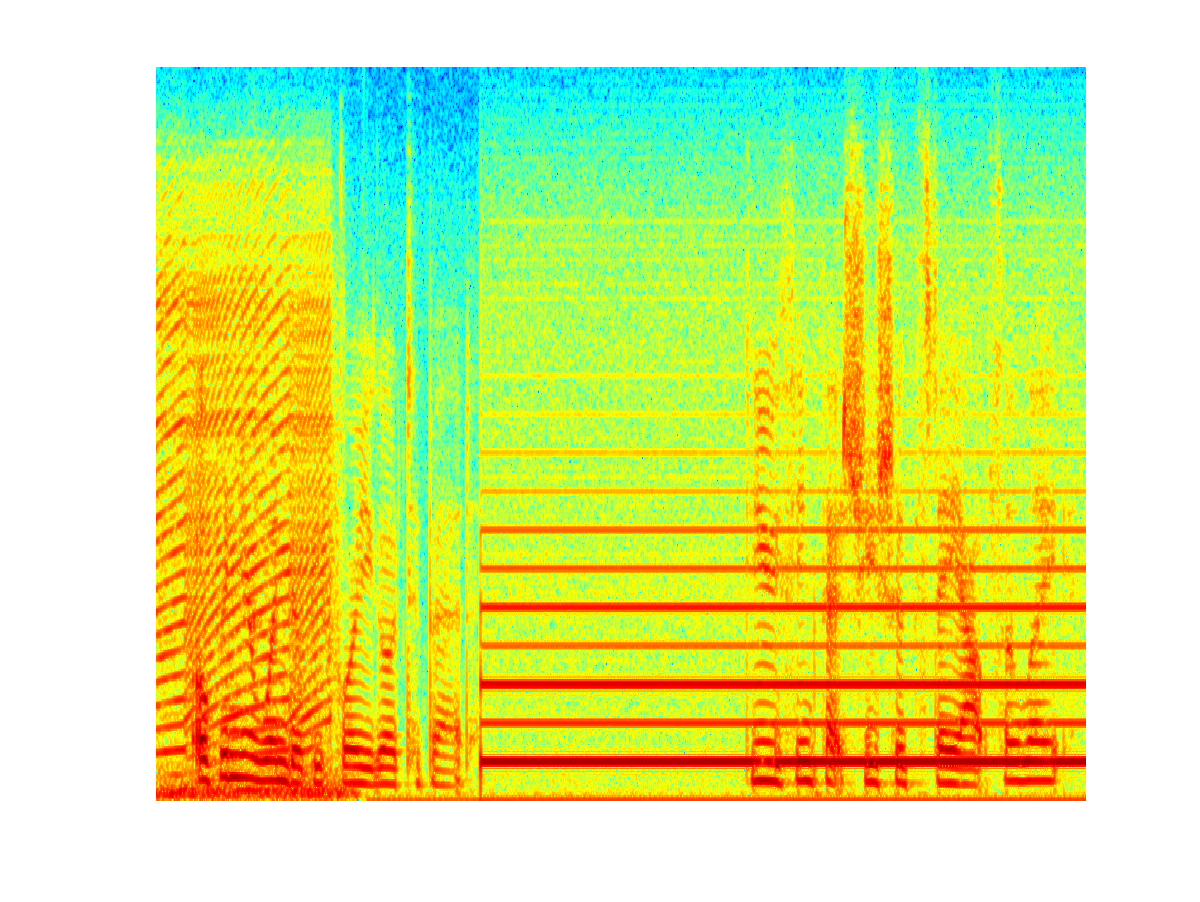}
    \caption{Signal corrupted with alarm noise at -3 dB SNR}
  \end{subfigure}

  \begin{subfigure}{\linewidth}
    \captionsetup{justification=centering,margin=0.1cm}
    \includegraphics[width=9cm,height=2.8cm,trim=80 60 0 0,clip=true]{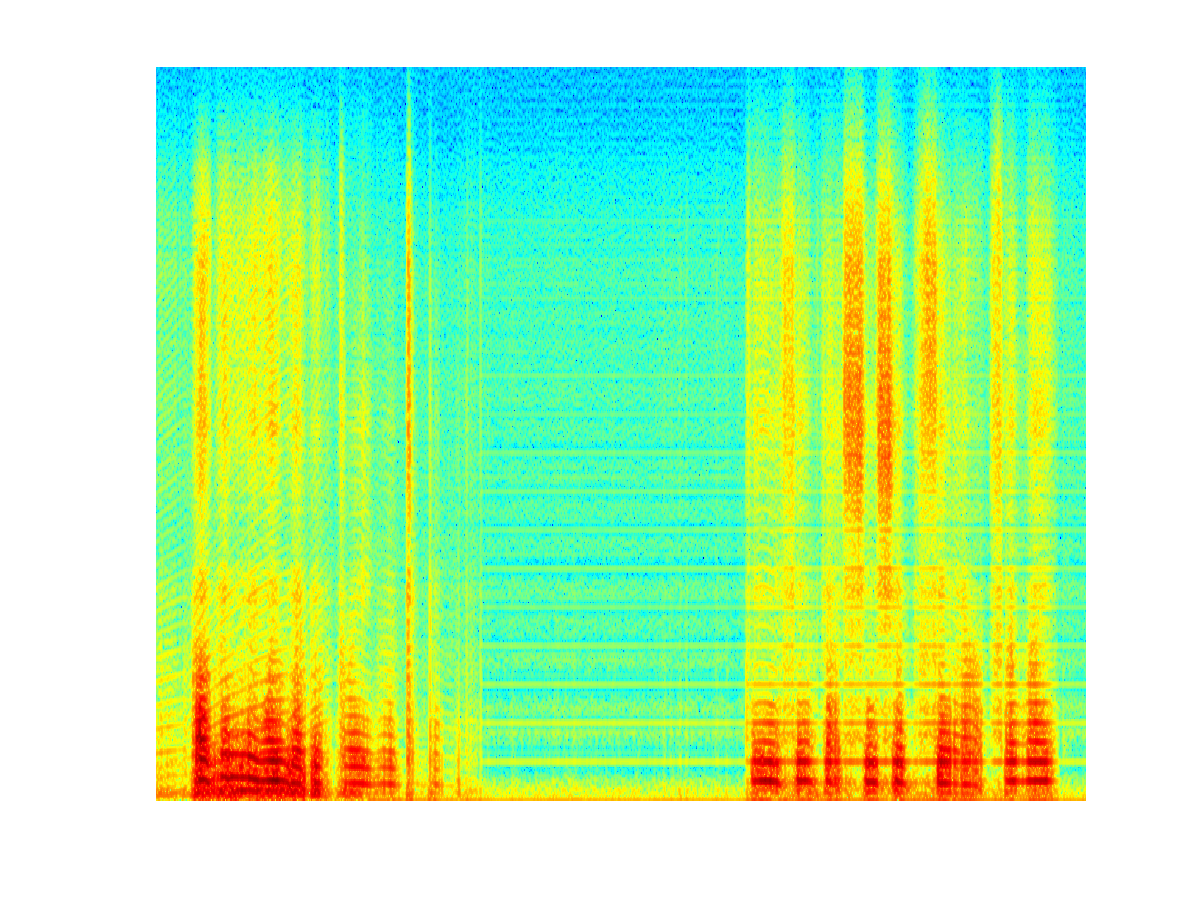}
    \caption{Signal denoised by a Single-Channel-DNN}
  \end{subfigure}

  \begin{subfigure}{\linewidth}
    \captionsetup{justification=centering,margin=0.1cm}
    \includegraphics[width=9cm,height=2.8cm,trim=80 60 0 0,clip=true]{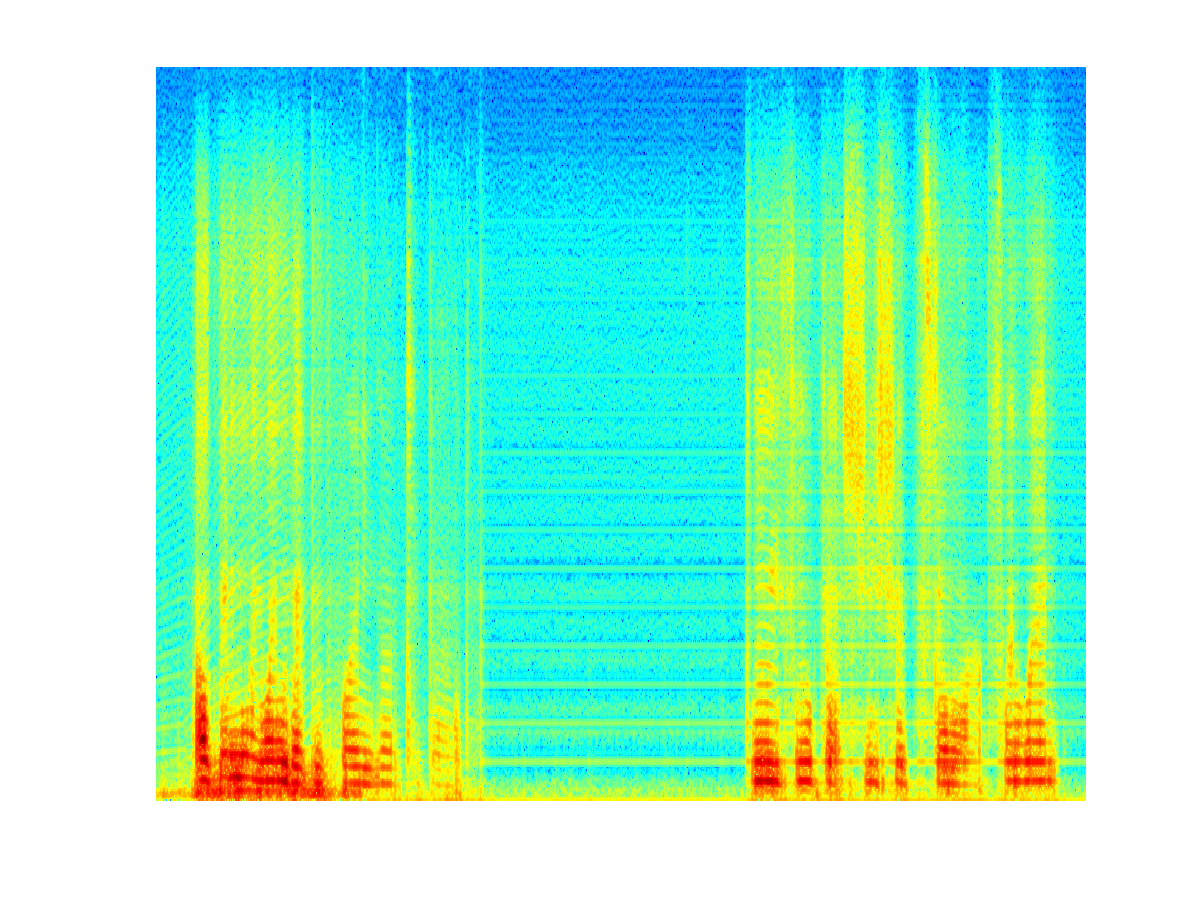}
    \caption{Signal denoised by a Single-Channel-BiLSTM}
  \end{subfigure}

  \begin{subfigure}{\linewidth}
    \captionsetup{justification=centering,margin=0.1cm}
    \includegraphics[width=9cm,height=2.8cm,trim=80 60 0 0,clip=true]{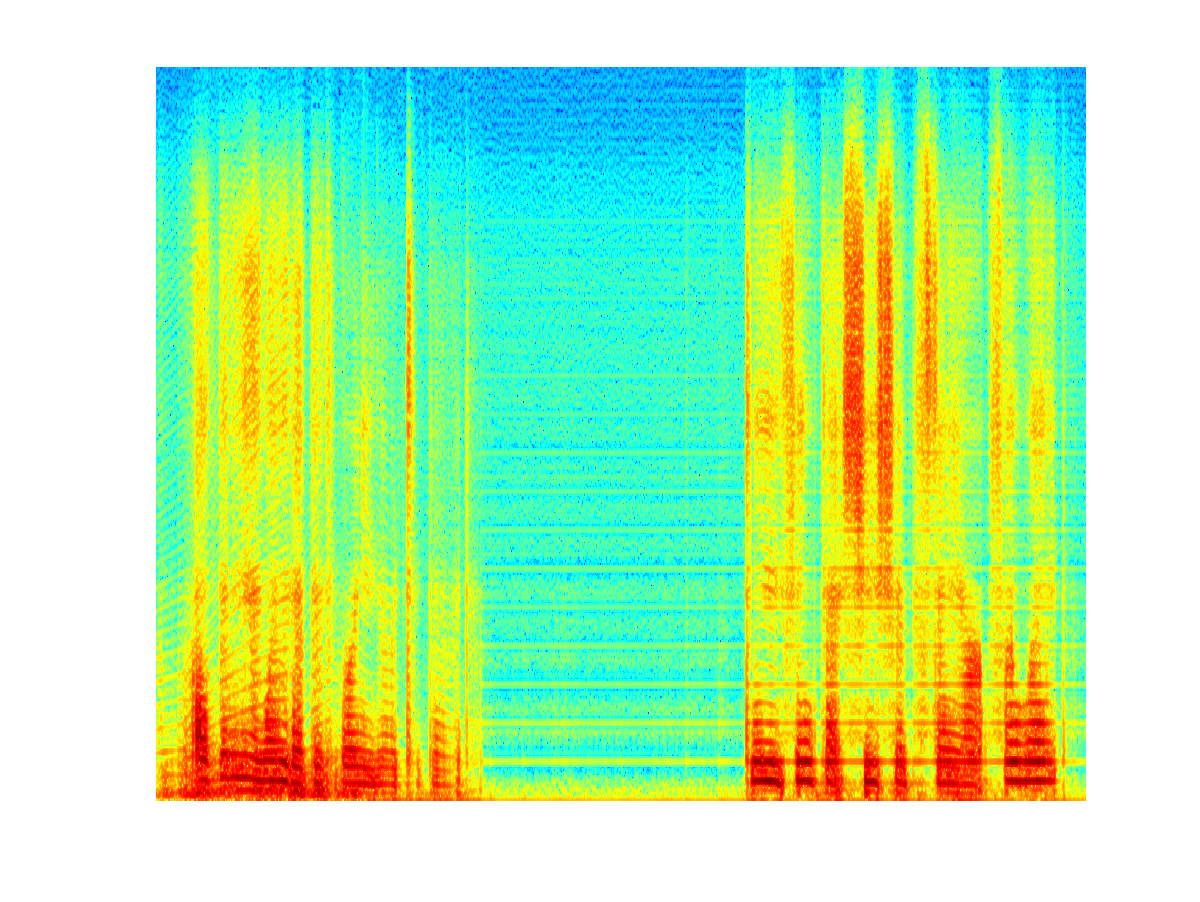}
    \caption{Signal denoised by a BiModal-BiLSTM}
  \end{subfigure}
\caption{Comparison of spectrograms of enhanced speech by different models}
\label{fig:spectrogram}
\vspace{-0.5cm}
\end{figure}

\subsection{Model Training}
All the models were trained on NVIDIA Tesla K20 GPUs using Theano \cite{bergstra+al:2010-scipy} and Mozi\footnote{https://github.com/hycis/Mozi.git}. We used Adam \cite{kingma2014adam} as the learning algorithm and Mean-Squared-Error as the objective to be minimized. We keep a 10\% of the training data as the validation set and stop training when the validation error has not improved over 5 epochs by at least 1\%. This ensures that none of the models over-fits to the training data. We normalise all audio input dimensions to have zero mean and unit variance, and scale the image pixel intensities to [0,1]. This pre-processing step is important to reduce co-variate shift across dimensions and to ensure that each dimension has equal signal intensity been passed to the network.

For the Single-Channel-DNN model, we used a window of 11 frames of the noisy spectrum for each output frame of the clean spectrum. For the BiLSTM models, each input time-step takes in 1 frame of speech and image. We also tried on windows of 3 to 7 frames for each input time-step, but we found that 1 frame worked the best. 

For the BiLSTM models, we unrolled the model with 21 time-steps, and trained with back-propagation through time \cite{werbos1990backpropagation}. We found that this gave a good balance between training time and model accuracy. Table \ref{table:cv-mse} shows the final Mean-Squared-Error (MSE) on the validation set. It can be seen that the proposed model has the least error, which indicates that the visual cues are helping in denoising the acoustic features.

\subsection{Results}
We use the Perceptual Evaluation of Speech Quality (PESQ) \cite{rix2001perceptual}, which has a high correlation with subjective evaluation scores, as our objective measure for evaluating the quality of denoised speech. Figure \ref{fig:pesq_scores} shows the average PESQ score of speech enhanced by different models on test utterances corrupted with seen noise (alarm and crowd) and unseen noise (traffic) at different SNRs. Table \ref{table:mean_pesq} shows the mean PESQ score across all speakers and all SNRs for the various models. We note that the mean PESQ scores are consistent with the MSE on the cross-validation set. The BiModal-BiLSTM performs best across all seen noises and SNRs but its performance is closer to Single-Channel-BiLSTM under the (unseen) traffic noise conditions. Both BiLSTM models significantly outperform the DNN model. Figure \ref{fig:spectrogram} shows the spectrogram of speech corrupted by alarm noise enhanced by different models. It can be seen that the noise is highly non-stationary and overlaps significantly with the speech spectral characteristics. All the models denoise reasonably well. This shows that visual information of lip movements indeed provide additional information in enhancing speech, and that a recurrent neural network is an effective model in learning this BiModal audio-visual information.
Since the information provided by the visual stream can only discriminate the manner of articulation, we initially suspected that most of the gains were coming from the suppression of noise in the silence frames. However, as can be seen from the spectrogram, the BiModal-BiLSTM also provides more details to the speech segments.

\section{Conclusions}\label{conclusion}
 Higher level information processing in human perception involves multi-sensory integration and modeling of long-term dependencies among the sensory data. Strategies involve integrating cues from multiple senses based on their reliability or Signal to Noise Ratio (SNR).  In this paper, motivated by the insights gleaned from human sensory perception, we have proposed a novel multi-modal hybrid deep neural network architecture. The model captures intermediate level representations of speech and images through a fully connected DNN and CNN respectively. The long term dependencies in the intermediate representation are modeled by a BiLSTM.  We validated the model on audio-visual speech enhancement, where the task is to estimate clean speech spectra from input noisy speech spectra and images of the corresponding lip region. It is expected that the hybrid model learns to adjust the importance of the audio and visual streams intrinsically based on the uncertainty in the audio stream. The hybrid model is trained jointly using the Backpropagation algorithm. We show that the proposed model achieves higher PESQ score on an average over a range of nonstationary noises and SNRs.  

  \eightpt
  \bibliographystyle{IEEEbib}

  \bibliography{mybib}

\end{document}